\documentclass[11pt]{article}
\usepackage{times}
\usepackage{latexsym}
\usepackage{amsmath}
\usepackage{amsfonts}
\usepackage{graphicx} 
\usepackage{url}
\usepackage{color}

\title{Neural Multi-task Learning in Automated Assessment}

\author{Ronan Cummins and Marek Rei \\
  ALTA Institute, \\
  The Computer Laboratory, \\
  Cambridge, CB3 0FD  \\
  {\tt firstname.surname@cl.cam.ac.uk}}

\date{}

\begin{document}

\maketitle

\begin{abstract}

Grammatical error detection and automated essay scoring are two tasks in the area of automated assessment. Traditionally these tasks have been treated independently with different machine learning models and features used for each task. In this paper, we develop a multi-task neural network model that jointly optimises for both tasks, and in particular we show that neural automated essay scoring can be significantly improved. We show that while the essay score provides little evidence to inform grammatical error detection, the essay score is highly influenced by error detection. 
  
\end{abstract}

\section{Introduction}

Automated assessment tools have been shown to be extremely valuable to second language learners. In particular, they enable learners to iteratively refine their writing in response to near instantaneous feedback. Grammatical error detection (GED) and automated essay scoring (AES) are two of the more prominent tasks in this area. 

Recent work \cite{Rei16} has modelled the task of GED as a binary sequence labelling task where a system aims to predict whether a particular word is errorful.\footnote{For missing words, the word following the place where missing word should go is labelled an error.} For example, a GED system would aim to detect the spelling and punctuation errors (in red) in the following sentence:

\begin{quote}
\emph{How low has President Obama gone to \textcolor{red}{tapp} my phones during the very sacred election process\textcolor{red}{.}}
\end{quote}

On the other hand, AES is a task that assigns a holistic quality score relating to the entire discourse. Typically, supervised machine learning approaches to these problems have focused on optimising only one of these objectives. However, it is likely that these two tasks are dependent on each other. For example, in an essay of poor quality, there may be a higher likelihood of grammatical errors throughout the essay. And conversely, the number and type of grammatical errors found is likely to be a useful indicator of essay quality. Therefore, we develop a multi-task learning neural network model that is trained jointly on both GED and AES in the domain of English as a second language (ESL) texts.

\section{Related Work}

There has been extensive work on both GED and AES in recent years. GED has often been treated in combination with the task of grammatical error correction (GEC) \cite{feliceYAYK14,RozovskayaR16} for which there has been several recent shared-tasks \cite{Conll13,Conll14}. However, there are situations where it is useful to treat GED as an isolated task. Recently, recurrent neural networks, specifically Bidirectional LSTMs (Long Short Term Memory), have been successfully applied \cite{Rei16, Rei16:2, Rei17} yielding state-of-the-art results. We build directly upon these systems in this work.

Commercially available AES systems have been available for some time and include PEG (Project Essay Grade)~\cite{page2003project}, e-Rater~\cite{attali2006automated}, and Intelligent Essay Assessor (IEA) \cite{landauer1998introduction}. In the academic arena, SVMs \cite{yannakoudakis2011}, Linear Regression \cite{phandi-chai-ng:2015:EMNLP}, and multi-task learning have all previously been applied to AES \cite{phandi-chai-ng:2015:EMNLP,cumminsZB16}. Although latterly the multi-task learning was adopted to tackled the problem of combining training data from different writing tasks and different scoring scales.

Deep neural network approaches to AES have also been developed \cite{AlikaniotisYR16,TaghipourN16,Zhao:2017}. Taghipour et al. \cite{TaghipourN16} studied a number of neural architectures for the AES task and determined that a bidirectional LSTM with mean pooling was the best performing single architecture on the ASAP (automated student assessment prize) dataset.\footnote{\url{https://www.kaggle.com/c/asap-aes}} However, all previous neural network approaches to AES were developed for text produced by native English learners and it is unclear if the results extend to ESL texts.

\section{Multi-task BiLSTM}

\begin{figure*}[!ht]
\begin{center}
$\begin{array}{c}
	\includegraphics[height=7cm,width=10cm]{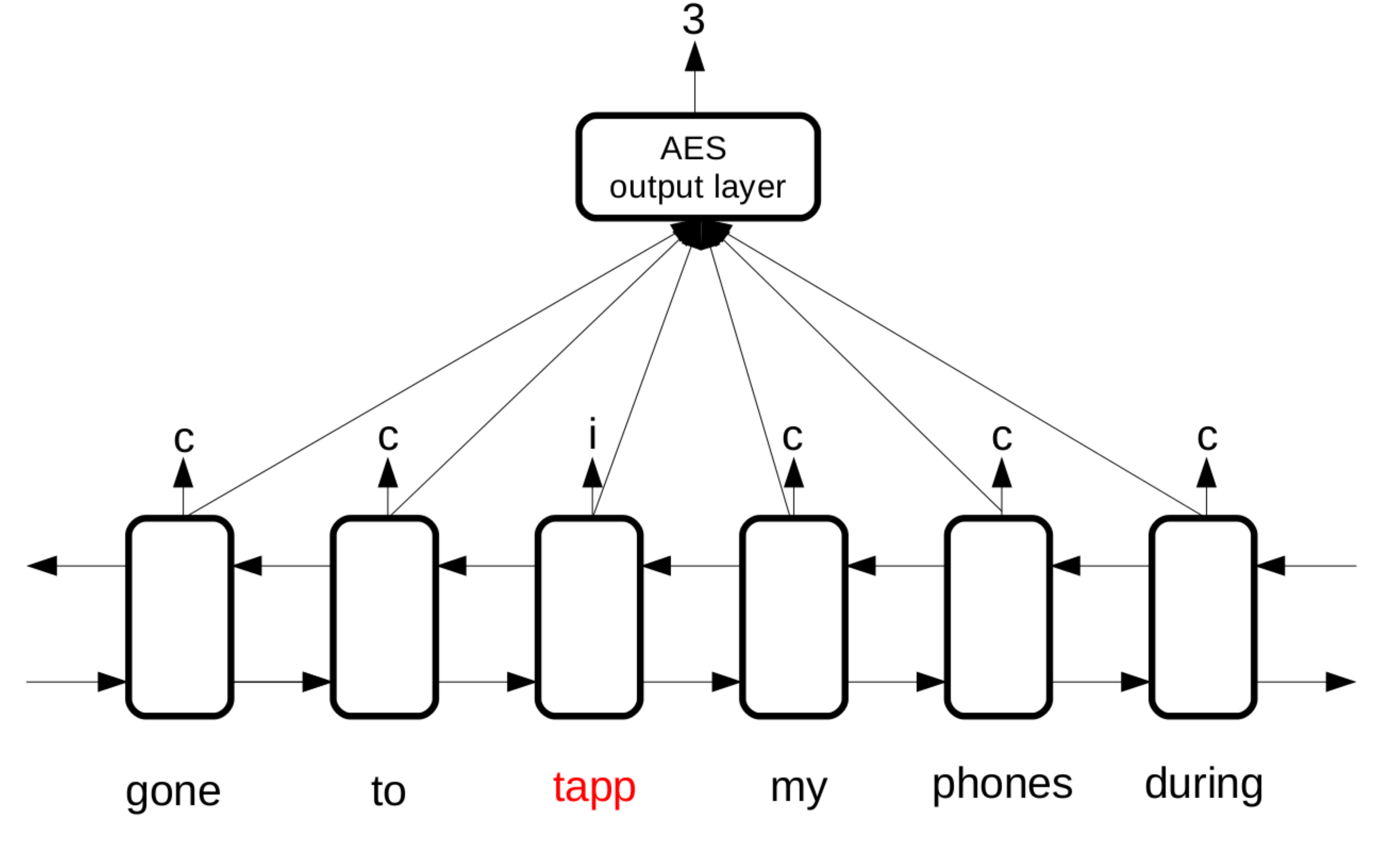} 	 	
\end{array}$
\caption{Bidirectional LSTM with multi-task objective of GED and AES for a low quality essay, where {\bf c} is correct, {\bf i} is incorrect, and {\bf 3} indicates the essay score predicted by the model. For simplicity, the semi-supervised language modelling component is not shown. \label{fig:bilstm}}
\end{center}
\end{figure*}

As a starting point, we used an existing bidirectional LSTM model for the error detection task \cite{Rei16,Rei17}. This model already contains an auxiliary semi-supervised objective, that of language modelling, which has been shown to improve detection performance. Although, this sequence labelling model was originally trained and tested on sentences, in this work we treat the entire essay as a sequence. Therefore, we augmented this model such that it also predicts an essay score after the entire sequence has been read in.

In particular, we modified the neural network (Figure~\ref{fig:bilstm}) such that the output of the hidden states in each direction are concatenated and averaged over all time-steps before being fed into the AES output layer. The AES output layer maps this concatenated vector to a single value, feeds this value into a sigmoid function and scales it appropriately. The scaling ensures that the model predicts a value between 1 and 20 (the range of scores for our dataset in Table~\ref{tab:transform}). The AES loss used is the squared error between the predicted score and the gold standard essay score. This AES architecture is the same as that identified in previous research \cite{TaghipourN16} as the best neural architecture for the AES task.

As a result, for each sequence (essay), the neural network optimises a combination of the error detection loss ($E_{ged}$) (cross-entropy), the language modelling loss ($E_{lm}$) (cross-entropy), and the essay scoring loss ($E_{aes}$) (mean squared error). The relationship between these loss functions is controlled by two hyperparameters, $\gamma_{_{lm}}$ and $\gamma_{_{aes}}$, as follows:

\begin{equation}
E = (1 - \gamma_{aes}) \cdot (E_{ged} + \gamma_{_{lm}} E_{lm}) + \gamma_{_{aes}} E_{aes}
\end{equation}
where $\gamma_{_{lm}}$ is set to its recommended setting of $0.1$ \cite{Rei17} for all subsequent experiments. The $\gamma_{_{aes}}$ parameter controls the importance given to the AES loss. The tuning of the $\gamma_{_{aes}}$ is explored in our experiments.

\section{Experiments}

The main aim of our experiments is to investigate how the multi-task loss function affects the performance of both the GED and AES tasks. 

\subsection{Dataset}

We use the public FCE dataset for our experiments.\footnote{\url{https://www.ilexir.co.uk/datasets/index.html}} The dataset contains short essays written by ESL learners for the First Certificate in English (FCE) (an upper intermediate level exam) \cite{yannakoudakis2011}. To our knowledge, this is the only public dataset that has annotations for both GED and AES on the same texts. The dataset contains 1244 scripts where each script contains two essays. After consultation with the curators of the dataset, we used the \emph{exam score} label in the FCE dataset mapped to the 20 point scale in Table~\ref{tab:transform} as the essay score.\footnote{Exam scores of 0 were removed} Our final training, development, and test set contains 2059, 198, and 194 essays respectively. 

\begin{table}[!ht]
\centering
\begin{tabular}{| l | l |}
\hline
Exam Score & Essay Score\\
\hline
1.1 & 1		\\
1.2 & 4		\\
1.3 & 8		\\
2.1 & 9		\\
2.2 & 10	\\
2.3 & 11	\\
3.1 & 12	\\
.. 	& ..	\\
..	& ..	\\
5.3 & 20	\\
\hline
\end{tabular}
\caption{Mapping of FCE \emph{\textless exam score \textgreater} to essay scores}
\label{tab:transform}
\end{table}

\subsection{Evaluation Metrics and Hyperparameter Settings}

The evaluation metric chosen for the GED task is $F_{0.5}$ which was the main metric used in the CoNLL-2014 shared task \cite{Conll14}. For the AES, task we used quadratic weighted kappa (QWK) which was the main metric used in the the ASAP Kaggle competition and many previous studies \cite{ChenH13}.

\begin{figure}[!ht]
\begin{center}
$\begin{array}{c}
	\includegraphics[height=5cm,width=6cm]{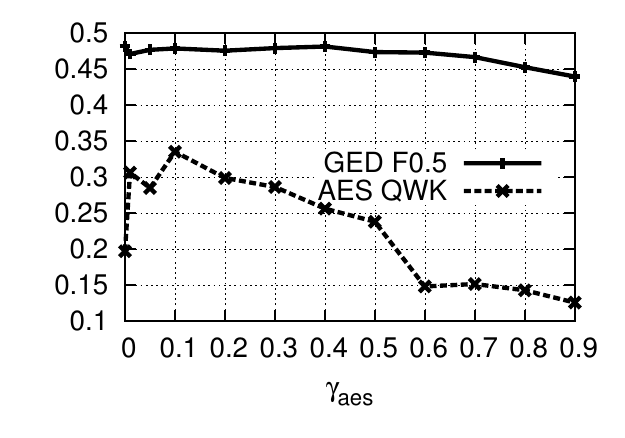} 	 	
\end{array}$
\caption{Tuning $\gamma_{_{aes}}$ for both GED and AES tasks on the development dataset. \label{fig:tuning}}
\end{center}
\end{figure}

There are a number of hyperparameters in our multi-task neural model. We use the publicly available Google News embeddings\footnote{Available at \url{https://code.google.com/archive/p/word2vec/}} (300 dimensions) to initialise the input layer. We used Adadelta optimisation with mini-batches. To prevent over-fitting we used \emph{early-stopping} where if the evaluation metric on the development set did not improve after seven epochs we stopped and reported the performance of the model that performed best on the development set. We tuned the AES cost parameter $\gamma_{_{aes}}$ on the development set from $0.0$ to $1.0$ using increments of $0.1$ as shown in Figure~\ref{fig:tuning}. 

\section{Results}

The optimal setting for the $\gamma_{_{aes}}$ hyperparameter on the development set for both tasks is shown in Figure~\ref{fig:tuning}.  The optimal setting of $\gamma_{_{aes}}$ is 0.4 for GED and 0.1 for AES. In particular, for GED the $\gamma_{_{aes}}$ parameter seems to make very little difference for a large part of the parameter space.

\subsection{Grammatical Error Detection}

Table~\ref{tab:ged} shows the performance of the neural multi-task model with and without the AES objective (+ $E_{aes}$) for the GED task. We see a small but insignificant increase in performance ($F_{0.5}$) when using the AES training objective. 

\begin{table}[!ht]
\centering
\begin{tabular}{| l | l | l | l | }
\hline
\multicolumn{4}{|c|}{Error Detection}						\\
\hline
Cost	 				& Precision & Recall & $F_{0.5}$			\\	
\hline
%baseline	& Sentence			& 0.591	& 0.204	&	0.429			\\
%+ lmcost	& Sentence			& 0.559	& 0.278	&	0.465			\\
%+ aescost	& Sentence			&   	& 		&					\\
%\hline
$E_{ged}$			& 0.492	& 0.251	&	0.413			\\
+ $E_{lm}$			& 0.588	& 0.221	&	0.442			\\
+ $E_{aes}$			& 0.543	& 0.265	&	0.449			\\
\hline
\end{tabular}
\caption{Performance of BiLSTM model on public-FCE test set for GED with multi-task training objectives (cumulatively) for entire essay contexts.}
\label{tab:ged}
\end{table}

\subsection{Automated Essay Scoring}

Table~\ref{tab:aes} shows the results of the neural multi-task model on the AES task with and without the GED objective (+ $E_{ged}$). In particular, we see that the semi-supervised language modelling objective ($E_{lm}$) improves the performance of the model. However, more strikingly we see that when the GED objective is added it increases the performance of our model on the AES task substantially. This is an encouraging result as it means that a neural model can utilise multiple signals in an end-to-end manner to improve AES. 

\begin{table}[!ht]
\centering
\begin{tabular}{| l | l | l |  }
\hline
\multicolumn{3}{|c|}{Automated Essay Scoring}		\\
\hline
Cost	 			& Spearman & QWK 	\\	
\hline
%nea-50		& Essay			& 0.416	&	0.352	\\
%nea-300		& Essay			& 0.417	&	0.373	\\
%\hline
$E_{aes}$		& 0.334	& 0.324		\\
+ $E_{lm}$		& 0.376	& 0.347		\\
+ $E_{ged}$		& 0.537*	& 0.459*		\\
\hline
\end{tabular}
\caption{Performance of BiLSTM AES task on public-FCE test set with and without multi-task training objectives (cumulatively) for entire essay context. * indicates that the results are statistically significant when compared to + $E_{lm}$}
\label{tab:aes}
\end{table}

\subsection{Third-Party Comparison of AES}

We now compare our best neural multi-task learning model on AES (Our Model) to an existing neural AES model\footnote{Available at \url{https://github.com/nusnlp/nea}} \cite{TaghipourN16} (labelled NEA). We ran NEA with both the 50 dimensional embeddings used in the original research (NEA-50) and the 300 dimensional embeddings used in this work (NEA-300), and the otherwise default settings with their code. To ensure that the comparison with the NEA system was fair, we successfully replicated the results of the original research \cite{TaghipourN16} on the ASAP dataset. We can see in Table~\ref{tab:third-party} that our multi-task model is substantially (and significantly) better than the NEA models. 

\begin{table}[!ht]
\centering
\begin{tabular}{| l | l | l | }
\hline
\multicolumn{3}{|c|}{Automated Essay Scoring}					\\
\hline
System	 			& Spearman & QWK		\\	
\hline
NEA-50				& 0.416	&	0.352	\\
NEA-300				& 0.417	&	0.373	\\
Our Model			& 0.537	& 	0.459	\\
\hline
\end{tabular}
\caption{Comparison of our multi-task model with third-party neural system for AES task.}
\label{tab:third-party}
\end{table}

\section{Discussion and Conclusion}

We have developed a neural multi-task system that jointly optimises the GED and AES tasks. Although, the GED task does not seem to benefit from incorporating information from the overall essay score, the AES task is substantially improved by the GED task. Our neural model allows multiple training signals to be combined in an end-to-end fashion.

%\section*{Acknowledgments}

%\bibliography{acl2017}
\bibliographystyle{plain}
\bibliography{bea2017}

\end{document}